\documentclass{llncs}
\usepackage{makeidx}  %
\usepackage{amsmath,amssymb} %
\usepackage[hidelinks]{hyperref}
\usepackage{url}
\usepackage{graphicx} %
\usepackage{caption}

\usepackage{booktabs}
\usepackage{makecell}
\usepackage{gensymb}
\usepackage{cite}
\usepackage{color,soul}

\usepackage[labelformat=simple]{subcaption}

\newtheorem{hyp}{Hypothesis}

\begin{document}
\frontmatter          %
\pagestyle{headings}  %
\addtocmark{Reducing the C-Space by SONN} %
\mainmatter              %

\title{Comparing SONN Types for Efficient Robot Motion Planning in the Configuration Space}

\titlerunning{Reducing the C-Space by SONN}  %

\author{Lea Steffen\inst{1} \and Tobias Weyer\inst{1} \and Katharina Glueck\inst{1}
 \and Stefan Ulbrich\inst{1} \and \\ Arne Roennau\inst{1} \and R\"udiger Dillmann\inst{1}}
\authorrunning{Lea Steffen et al.} %
\institute{FZI Research Center for Information Technology, 76131 Karlsruhe, Germany,\\
\email{steffen@fzi.de}}

\maketitle              %

\begin{abstract}
	Motion planning in the configuration space (C-space) induces benefits, such as smooth trajectories. It becomes more complex as the degrees of freedom (DOF) increase. This is due to the direct relation between the dimensionality of the search space and the DOF.
	Self-organizing neural networks (SONN) and their famous candidate, the Self-Organizing Map, have been proven to be useful tools for C-space reduction while preserving its underlying topology, as presented in \cite{Steffen2021_dimreduction}. 
	In this work, we extend our previous study with additional models and adapt the approach from human motion data towards robots' kinematics.
	The evaluation includes the best performant models from \cite{Steffen2021_dimreduction} and three additional SONN architectures, representing the consequent continuation of this previous work. Generated Trajectories, planned with the different SONN models, were successfully tested in a robot simulation.
\end{abstract}

\section{Introduction} \label{sec:introduction}
The core problem of robot motion planning is how to transition a robot arm from one pose to another. %
A probate algorithm for trajectory planning is the Wavefront algorithm, a breadth-first search \cite{Naumov2015}. Its execution in the configuration space (C-space) is advantageous, however, it struggles due to the high dimensions. For a neural implementation within the C-space, neurons represent discrete configurations while synapses are applied for path planning. The impressive success of self-organizing neural networks (SONNs) is based on their simple definition, effective implementation and its features regarding clustering and visualization \cite{Cottrell2016}. Hence, SONNs are a powerful and versatile tool capable to reduce the complexity of neural motion control through pruning unnecessary neurons and synapses with simultaneous preservation of the topology.
SONNs are still in an early stage, however, several approaches of self-organization applied to %
motion control \cite{Saxon1990, Martinetz1990, Barreto2003, Fontinele2016} have been introduced.
Already at the beginning of the 90s, there were attempts to simplify robot movements by means of SONNs. Firstly, in \cite{Martinetz1990}, a hierarchical architecture has been proposed, embodying many SOMs learns visuomotor coordination for a robot arm with 5 DOF. Secondly, in \cite{Saxon1990} a representation of the joint and Cartesian positions of a robot arm has been represented by one single SOM. A performance comparison of several learning algorithms being able to compute the inverse kinematics (IK) of a redundant robotic arm has been presented in \cite{Fontinele2016}.
A survey about SOM-based approaches for controlling robotic arms is given in \cite{Barreto2003}. The authors cover a broad spectrum from obstacle avoidance to hand-eye coordination and computation of IK. A general overview about neural networks for robot motion control, not limited to self-organization, is provided in \cite{Prabhu1996}. \\
In our previous paper \cite{Steffen2021_dimreduction}, we presented a framework to reduce the dimension of a C-space with SONNs and a solid proof of concept that this reduced C-space is capable to support planning of trajectories. In this paper we present an in depth evaluation about different SONN models, which allows to exclude some former models and introduce new ones, as presented in \autoref{sec:selected_SONN}. Based on the findings in \cite{Steffen2021_dimreduction}, we also formulate hypotheses \ref{hyp:1}-\ref{hyp:4} in \autoref{sec:methodology} and provide an analysis of them in \ref{sec:evaluation}. Lastly, we transferred the approach from 7 DOF human motions to 6 DOF robot trajectories as training data.

\section{Self-Organizing Neural Networks} \label{sec:sonn}
The self-organizing map (SOM) \cite{Kohonen1982} represents an unsupervised learning model. Thus, a machine learning technique is applied to detect structural information and patterns in unlabeled data. %
SOMs create a discretized representation of the input space in a lower dimension, thereby input signals are mapped so that similar stimuli are close to each other, i.e. stimulate neurons that are neighbors \cite{Kohonen2013}. 
The input layer is fully connected with the neurons in the competitive layer of the map, where learning takes place. 
The map's neurons are connected in an inhibitory mode, therefore new input hampers the neurons activity. For learning, an input vector $ x = [x_1...x_n]^T \in \mathbb{R}^n $ is presented in every step via the input layer to each neuron within the map. The neuron whose weight vector is most similar to the input signal is considered the winning best matching unit (BMU). It is common, but not mandatory, to use the Euclidean distance as a similarity metric \cite{Kohonen1982}. For the BMU and in a less pronounced form, its neighboring neurons, the weights are shifted towards the input signal. With this strategy, the network slowly approximates the underlying topology of the input data.
The SOM proved to be a sophisticated method for data exploration and has been gradually extended and refined in different directions. However, the original SOM does not provide a complete topology preservation as it is folded in the multidimensional space due to the 'topological mismatch', as stated in \cite{Martinetz1994}. Hence, all adjacent points in the reduced C-space should also be adjacent in the original C-space but not all adjacent points of the original Space are necessarily adjacent in the reduced C-space. Consequently, research has been focused on overcoming this issue over the last decades \cite{Villmann1997}. \\
Besides the SOM, a fundamental representative of topology preserving self-organizing learning algorithms is the Neural Gas (NG) \cite{Martinetz1991}. 
As NGs are unstructured and SOMs have a rigid structure their learning processes differ.
While a winner-takes-all approach is applied for the SOM, for NGs rely in extended competitive learning.
This means that the neurons, adapted to the input, besides the BMU, are the neurons whose weight vector is most similar to that of the BMU.
A prerequisite for using SOMs and NGs is the required initial definition of the number of neurons. As it is not always predictable which grid size, and in the case of SOMs also its dimension, provides the optimal clustering, this is a challenging problem. Consequently, a method for detecting the required grid size during learning, named Growing Neural Gas (GNG) \cite{Fritzke1995}, has been proposed .
With regard to SOMs handling temporal characteristics, there are several extensions that are roughly divided into 5 classes in literature \cite{Hammer2005, VanHulle2012} \textit{(1)} fixed-length windows \cite{Martinetz1993, Simon2003}, 
\textit{(2)} specific sequence metrics \cite{Kohonen1997, Somervuo2004},
\textit{(3)} statistical modeling incorporating appropriate generative models for sequences \cite{Bishop1997, Tino2004},
\textit{(4)} mapping of temporal dependencies to spatial correlation \cite{Euliano1999, Schulz2004, Wiemer2003}, 
\textit{(5a)} SOMs with supplementary feedback connections to enable recurrent processing of time signals through integrating a neurons past activations like the Temporal Kohonen Map \cite{Chappell1993} and the recurrent SOM (RSOM) \cite{Koskela1998}, \textit{(5b)} SOMs allowing integration of past states of other neurons as the Recursive SOM (RecSOM) \cite{Voegtlin2001}, the SOM for structured data (SOMSD) \cite{Hagenbuchner2003}, the MGNG \cite{Andreakis2009} ,the Merge SOM (MSOM) \cite{Strickert2005} and its extensions the $\gamma$-SOM \cite{Estevez2009} (see \autoref{fig:gamma_som}). A respective extension of the GNG was introduced in \cite{Estevez2013} as the $\gamma$-GNG.\\
A quite recent development, an extension of the GNG, taking a new approach, is the Segment GNG (SGNG) \cite{Vergara2016, Vergara2017} (see \autoref{fig:sgng}). This model differs strongly from other models because here, parts of the trajectory serve as input instead of joint angles. 
Overviews about SOM models are given in \cite{VanHulle2012, Hammer2005, Miljkovic2017}.

\section{Methodology} \label{sec:methodology}
\begin{figure}[h!]
	\centering
		\includegraphics[height=3cm]{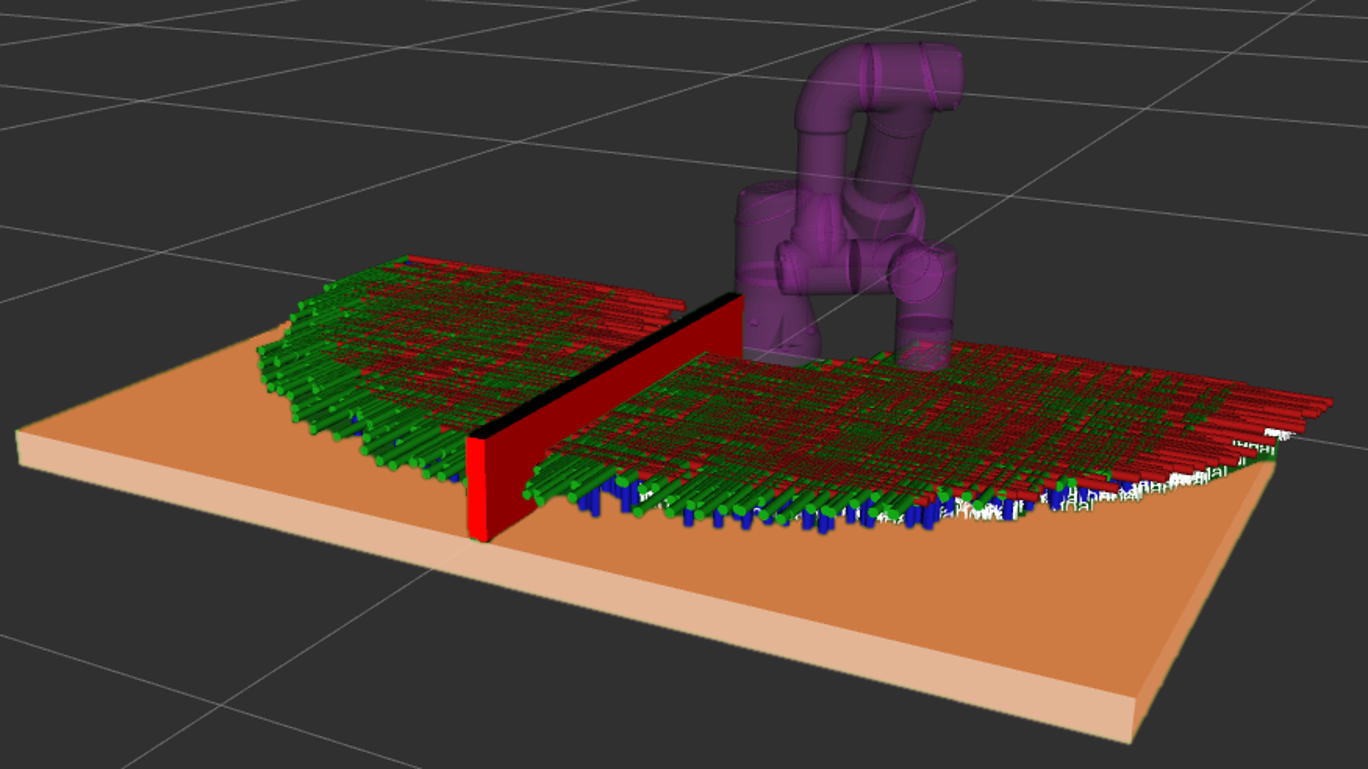}
	\caption{Experimental setup, a UR3 mounted on a table with a barrier. Coordinate systems of start and target positions for every training trajectory are spread 5 cm over the table.}
	\label{fig:ur_mounted}
\end{figure}

\begin{center}
	\centering
	\captionsetup{type=figure}
	\includegraphics[width=.95\textwidth]{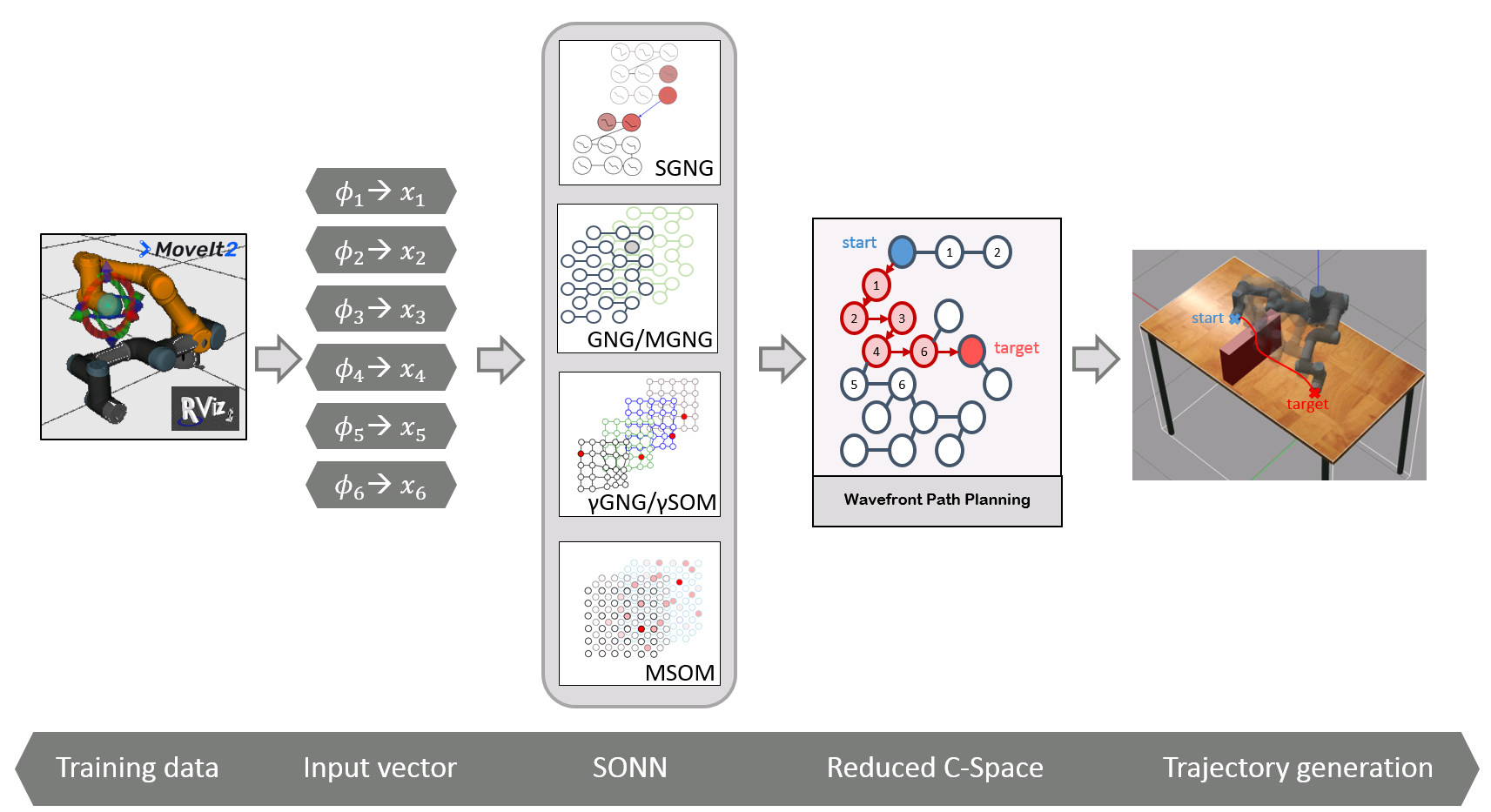}
	\captionof{figure}{A 6D input vector is derived from simulated UR3 trajectories. C-space dimensionality reduction is done via a selection of self-organized neural networks.
		Subsequently, motion planning is done in a reduced space by the Wavefront algorithm.}
	\label{fig:konzeptskizze}
\end{center}%
Robotic motion planning is very robust in the high dimensional C-space but requires expensive computations and enormous amounts of memory. Because robotic arm movements are highly redundant, similar motions can be clustered to reduce the C-space in \cite{Steffen2021_dimreduction}. In this work we build on these results and expand our previous study towards more potent models. Additionally, we transfer the approach from human motion data to a robot's kinematic. The most important novelties are, firstly, the implementation of new SONN architectures as a consequence of the evaluation results in \cite{Steffen2021_dimreduction}. This yields to better coverage. Secondly, the training data, previously consisting of human trajectories from the Master Motor Map reference model \cite{Terlemez2015}, was replaced by real robot trajectories as visualized in \autoref{fig:ur_mounted}. 
In addition, the different SONN types have been analyzed. Therefore, several hypotheses can be formulated, based on results of \cite{Steffen2021_dimreduction} and findings in literature~\cite{Andreakis2009, Estevez2009, Estevez2013, Vergara2016, Vergara2017}:
	\begin{hyp} \label{hyp:1}
		Densely sampled input sequences cause clustering as the same BMU is chosen for many iterations. Thus, sparser sampled data should be preferred.
	\end{hyp}

	\begin{hyp} \label{hyp:2}
		SONN models considering temporal context learn a better path preservation than versions which do not.
	\end{hyp}

	\begin{hyp} \label{hyp:3}
		The reason why GNG based models produce shorter paths in comparison to SOM based models is that GNGs establish many long connections which lead to shorter	paths but also to large joint angle jumps.
	\end{hyp}
	\begin{hyp} \label{hyp:4}
		GNG-based models achieve a better coverage of the input space compared to SOM-based models.
	\end{hyp}
In \autoref{sec:qualitative_analysis} a qualitative and in \autoref{sec:parameter_setting}, \ref{sec:sonn_comparison} \& \ref{sec:path_analysis}	a quantitative study is presented to verify the formulated hypotheses.
In \autoref{fig:konzeptskizze}, the conceptual sketch, including the new models and additional changes, are shown. Trajectories of a simulated UR3 are used to generate a 6D input vector, depicted at the left side of the graphic. The main module, applied to reduce the dimensionality of the C-space is shown in the center. It consists of 1 of 6 alternative SONN, as described in \autoref{sec:selected_SONN}. Lastly, the Wavefront algorithm can be performed in the reduced C-space to generate new trajectories. 

\subsection{Model selection} \label{sec:selected_SONN}
\begin{figure}[hbt!]
	\centering
	\begin{subfigure}[b]{0.49\textwidth}
		\centering
		\includegraphics[height=4.5cm]{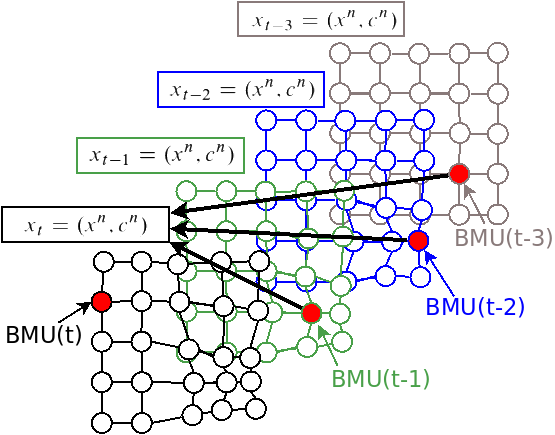}
		\caption{$\gamma$-SOM}
		\label{fig:gamma_som}
	\end{subfigure}
	\hfill
	\begin{subfigure}[b]{0.49\textwidth}
		\centering
		\includegraphics[height=4.5cm]{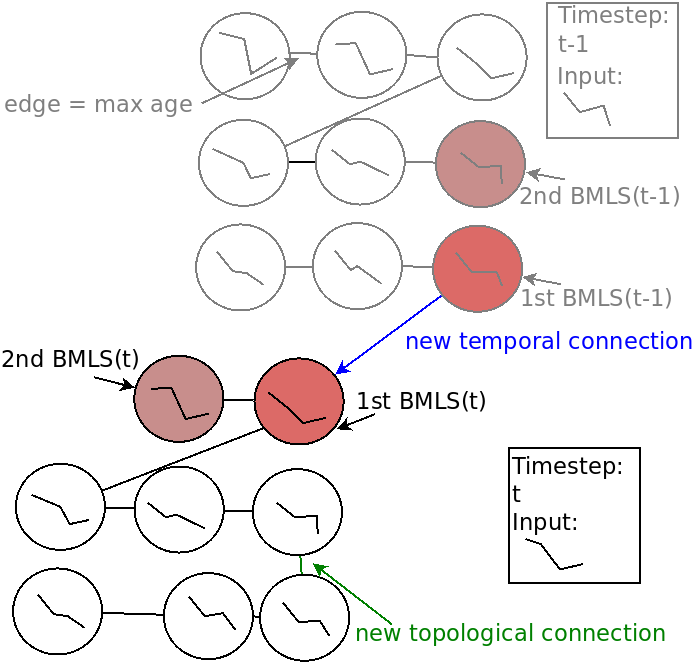}
		\caption{SGNG}
		\label{fig:sgng}
	\end{subfigure}
	\caption{\protect\subref{fig:gamma_som} A SGNG's learning phase. After the BMLS for $t$ is selected, a temporal connection is created between the current and previous BMLS. A topological connection, created between the 1st and 2nd BMLS at $t-1$, is visible at $t$. }
	\label{fig:som_models}
\end{figure}

The reduced space must be a topology and path preserving map to be suited for path planning. A graph G forms a topology preserving map of a manifold M, if the mapping from M to G as well as the inverse mapping from G to M is neighborhood preserving. Only then, G forms a map on which adjacent locations, neurons in the context of SONNs, correspond to features neighboring on M and vice versa \cite{Martinetz1994}. In other words, only with this definition of topology preservation it is possible to reduce the dimensionality of M. In this context, M is the C-space of the robot. And in addition, a topological path in G, the reduced space, has to be found and eventually re-transformed to a useful path in M. SOMs try to form a neighborhood preserving mapping from G to M, but not necessarily a neighborhood preserving mapping from M to G. Consequently, SOMs cannot theoretically provide a complete path preservation. 
However, it is very interesting from a scientific point of view to test these more rigid traditional SONN as well for the comparison of the models.\\
The network selection is based on the evaluation results from \cite{Steffen2021_dimreduction}. As listed in \autoref{tab:evaluation:result_first_paper}, it has been shown that path planning is not feasible in a reduced C-space generated with the NG and MNG. The reason is poor coverage of the input space and the lack of structure of these models. Thus, we did not continue both types in our study. The SOM and MSOM have deficiencies in regard to the coverage of the input space and they often generate paths with detours.
Overall the best coverage was achieved with the growing versions, GNG and MGNG. 
\begin{table}[h!]
	\begin{center}
		{\def\arraystretch{2}\tabcolsep=5pt
			\begin{tabular}{l c c c c} 
				\toprule
				& \textbf{C-Space Coverage} & \textbf{Structure}  & \textbf{QE}\\
				\midrule
				\textbf{SOM / MSOM} & \makecell{incomplete, but small \\ joint angle \\distances well represented} & \makecell{neighborhood of \\ neurons can be used \\ for path planning}  & high \\ 
				\textbf{NG/MNG } & incomplete & none  & very high \\ 
				\textbf{GNG/MGNG } & \makecell{good (depending on the \\amount of input data)} & \makecell{synapses can be used \\for path planning} & low \\ 
				\bottomrule
			\end{tabular}
		} %
		\caption{The table depicts the evaluation results from \cite{Steffen2021_dimreduction}. As only small differences were found between the two basic networks, the SOM and the NG, and their respective merging versions, the MSOM and MNG, they are each listed in one category. The two growing version, GNG and MGNG, were also summarized due to great similarities, but there are certain differences in terms of their properties. The quantization error is abbreviated as QE.}
		\label{tab:evaluation:result_first_paper}
	\end{center}
\end{table} 
As the SOM and MSOM performed quite similarly, however, the MSOM was slightly superior, the MSOM was retained and the SOM was excluded. Due to their surprisingly well coverage of the input space and as they have shown some different characteristics, both the GNG and the MGNG were retained. \\
For an additional model, a solution that implements the memory structure very well is required. 
The $\gamma$-SOM model \cite{Estevez2009}, as it belongs to class \textit{(5a)} of \autoref{sec:sonn}, is designed for temporal sequence processing. It extends the original SOM algorithm by $n$ context descriptors, referred to as the \textit{$\gamma$-memory} with depth $n$, a short term memory structure. With the $\gamma$-SOM, it can be defined how many steps back are memorized by means of the memory depth as shown in \autoref{fig:gamma_som}. If the memory depth is set to a single stage with $n=1$, the $\gamma$-SOM is identical to the MSOM \cite{Strickert2005}. The authors of \cite{Estevez2009} claim that increasing depth does not cause a lost in resolution. A respective addition of a context vector including a memory function with adjustable depth also exists for the GNG. The $\gamma$-GNG \cite{Estevez2013} combines the network property of growing models to adapt the number of neurons with the advantages of a short term memory as context. \\
As the Segment GNG (SGNG) \cite{Vergara2016}, an extension of the GNG, is said to have very good adaptation properties and a good temporary context representation, we included it into our study. The SGNG approximates small parts of the trajectories by segments, which sounds perfect for the purpose of trajectory learning in SONNs.
The first aspect that distinguishes this model from the GNG, is that the network's basic unit is altered from a node to a linear segment between two nodes. Its second distinguishing feature is the addition of temporal connections between the winning neurons of two succeeding steps.
This method is fundamentally different from all the previous ones as it does not consider the joint angles but instead partial trajectories, representing segments, as basic units of quantization.
Hereby, the weight vector is irrelevant and the main unit is a segment. The segment quantizes portions of a trajectory,  by increasing the number of included past samples and evaluate the obtained portions by determining the linearity. 
Consequently, there is no BMU in this approach, but instead a Best Matching Linear Segment (BMLS). 
A training sample consists of a trajectory portion ${\varphi}^{t- \tau} _{i}$. To determine its BMLS, a distance measure considering two characteristics of all segments $S^i$ is applied. Firstly, $d_{close}$, the spatial closeness between each segment $S^i$ and the training sample ${\varphi}^{t- \tau} _{i}$ is calculated, by means of their respective midpoints. Secondly, $d_{parallel}$ the degree of parallelism between $S^i$ and the line joining the trajectory portion's extreme points ($\phi_{t} - \tau \phi_{t}$) is determined through its cosine similarity. 
After selection, a temporal connection is established between the current and previous BMLS, as visualized in \autoref{fig:sgng}. Additionally, a topological connection is established between the 1st and 2nd BMLS. In \autoref{fig:sgng}, the topological connection created at time step $t-1$ is visible at time step $t$.
A detailed description of the structure and learning properties of SGNG is given in \cite{Vergara2016, Vergara2017}.

\subsection{Connection reduction} \label{sec:connection_recudtion}
In some rare cases connections cause jumps between successive joint angles due to their length. 
An intuitive solution to control the length of the connections and thereby avoid jumps, is to delete connections whose distance is above the threshold of 10\degree.
In the course of this, the \textit{Chebyshev distance}, the largest difference between the joint angles of two connected neurons, was used for distance calculation. Before the deletion of a synapse, it must be assured that the two respective neurons are still connected to each other indirectly. Otherwise, the reduced C-space would fragment. To check this condition, a copy of the connections is created, the connection is deleted and a wavefront is sent through the network. If a path is still found, the copied reduced connections are chosen.

\section{Evaluation} \label{sec:evaluation}
The hypotheses postulated in \autoref{sec:methodology}, represent the foundation of the following qualitative (\autoref{sec:qualitative_analysis}) and quantitative (\autoref{sec:parameter_setting}, \ref{sec:sonn_comparison} \& \ref{sec:path_analysis}) analysis.

\subsection{Qualitative Analysis} \label{sec:qualitative_analysis}
To evaluate the learning behavior of the SONN models, a qualitative visual analysis was conducted with a 3D visualization. Thus, only the two shoulder joints and the elbow joint were considered. Furthermore, only 15 random training trajectories were used.
\begin{figure}[hbt!]
	
	\begin{subfigure}[b]{0.49\textwidth}
		\centering
		\includegraphics[height=3.5cm]{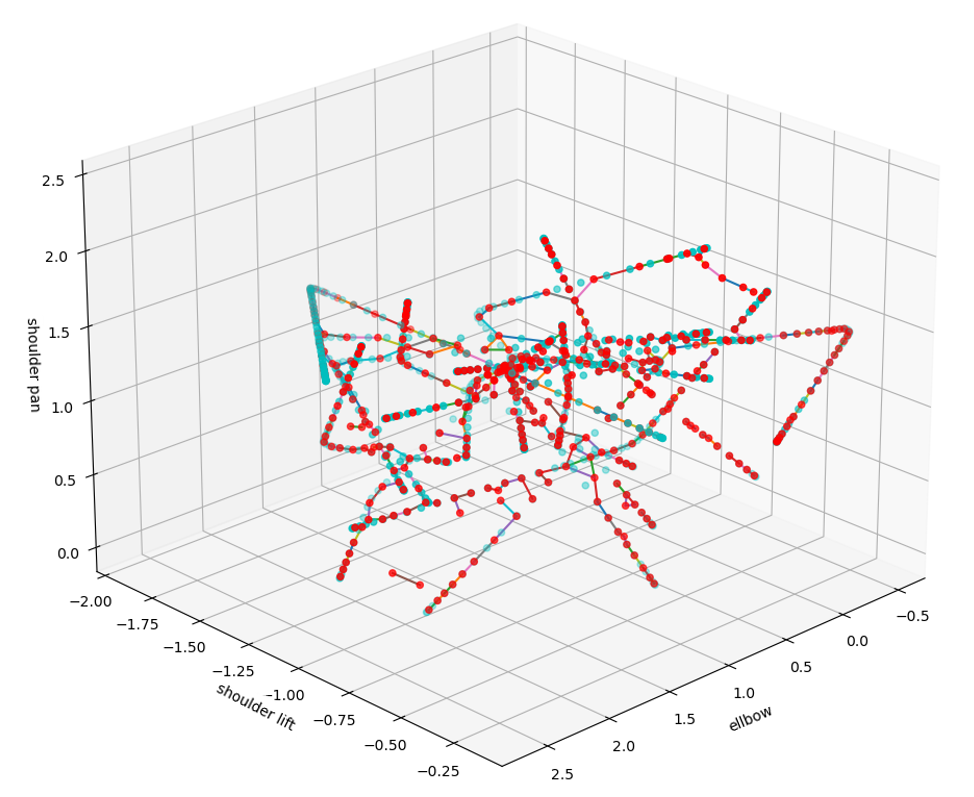}
		\caption{sparse}
		\label{subfig:original}
	\end{subfigure}
	\hfill
	\begin{subfigure}[b]{0.49\textwidth}
		\centering
		\includegraphics[height=3.5cm]{figures_paper/GNG_interpol4_steps70000_cut}
		\caption{dense}
		\label{subfig:inperpolated}
	\end{subfigure}
	\caption{
		GNG output spaces, visualized for three of the six joint angles, trained with original sparse vs. interpolated dense input data. Red scatter dots represent the learned weight
		vectors of the neurons. Lines between red dots indicate learned connections between neurons. The blue scatter dots represent input samples.}
	\label{fig:connections}
\end{figure}
A comparison between sparsely sampled and densely sampled input data, to verify hypothesis~\ref{hyp:1}, was carried out. Through repetitive interpolation (4x) the sample number was increased by factor 16.
To represent the results of the comparison between sparse and dense input, the GNG model was chosen, but all other SONN types showed a very similar behavior.
The output space of SONNs can be visualized by interpreting the neurons’ weights as coordinates. The 3D input space results in weight vectors with three elements which can be displayed in a 3D graph.
In \autoref{subfig:original}, the output space of a GNG with sparsely sampled input and in \autoref{subfig:inperpolated}, with densely sampled input is shown. Both variants achieve a good quantization of the input space and the connections between the neurons reproduce a good topological estimation of the input trajectories.
However, some topological connections between the neurons trained with sparsely sampled input trajectories appear to be broken at several positions. This leads to a splitting of the graph structure, causing problems for path planning. In contrast, the topological connectivity in the output space of densely sampled input trajectories corresponds almost exactly to the input space topology. A possible explanation is that splits occur with sparse input if neurons partially cover the input samples exactly. Hence, for two samples next to each other, these neurons will keep their positions in subsequent learning iterations while they also keep the same second BMU. Subsequently, only the connection to the constantly same second BMU is refreshed while the other connections to other neurons age and are eventually deleted. This effect cannot occur if the input space is densely sampled since there will always be enough samples around the neurons to generate changing BMU and second BMU pairs. Thus, a reasonable ratio between the number of input samples and the number of neurons is crucial. The graph splits if not enough neurons exist compared to the input samples, but trajectories cannot be properly approximated with too many neurons.
 However, any clustering of neurons for denser sampled input data, as
postulated in hypothesis~\ref{hyp:1}, could not be observed. On the contrary, denser sampled input data seems to prevent splitting of the topological structure.

\begin{figure}[h!]
	\centering
	\begin{subfigure}[b]{0.3\textwidth}
		\centering
		\scalebox{1}[1]{
			\includegraphics[width=\textwidth]{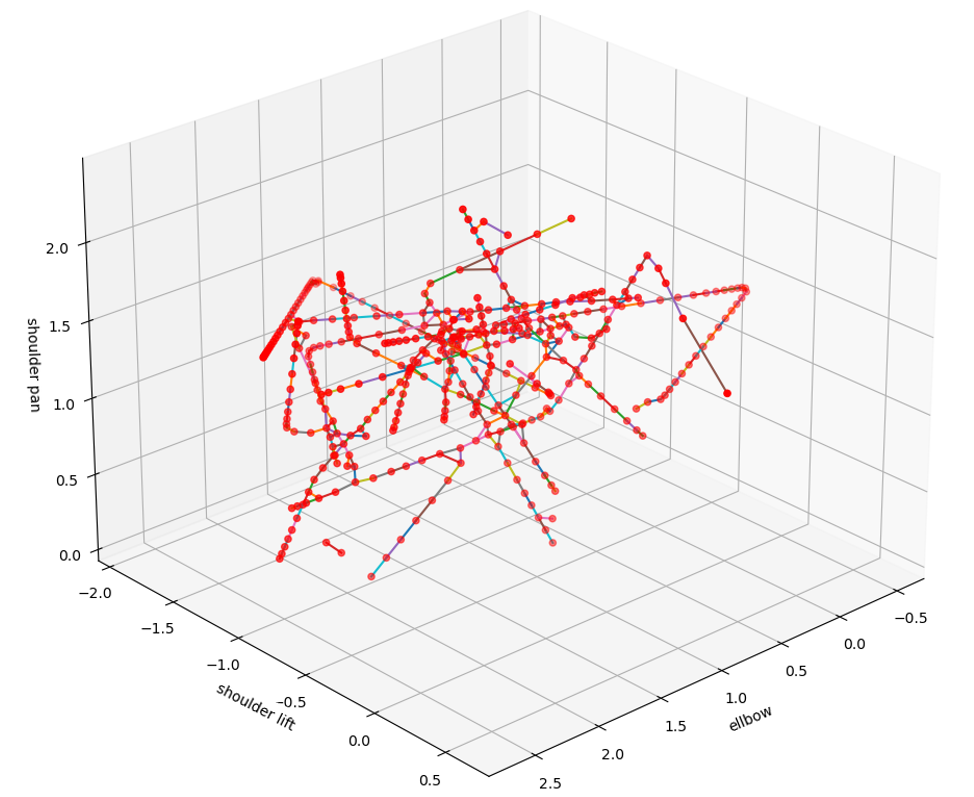}}
		\caption{\centering GNG}
	\end{subfigure}
	\begin{subfigure}[b]{0.3\textwidth}
		\centering
		\scalebox{1}[0.95]{\includegraphics[width=\textwidth]{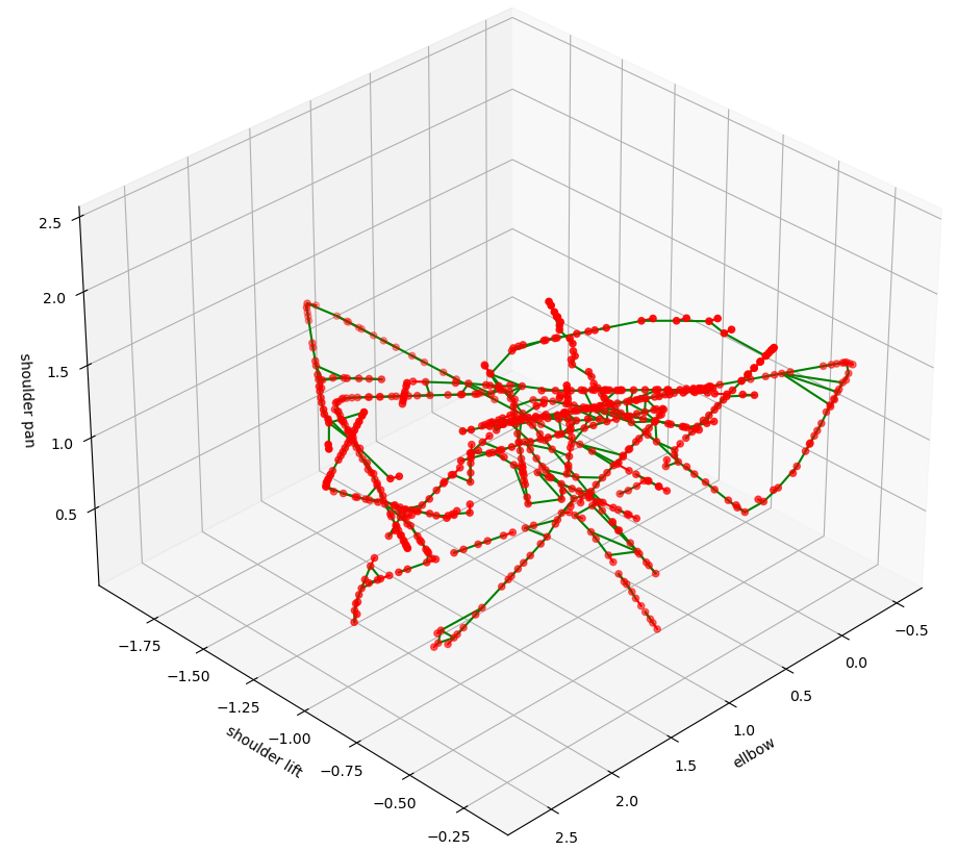}}
		\caption{\centering SGNG}
	\end{subfigure}
	\begin{subfigure}[b]{0.3\textwidth}
		\centering
		\scalebox{1}[0.95]{\includegraphics[width=\textwidth]{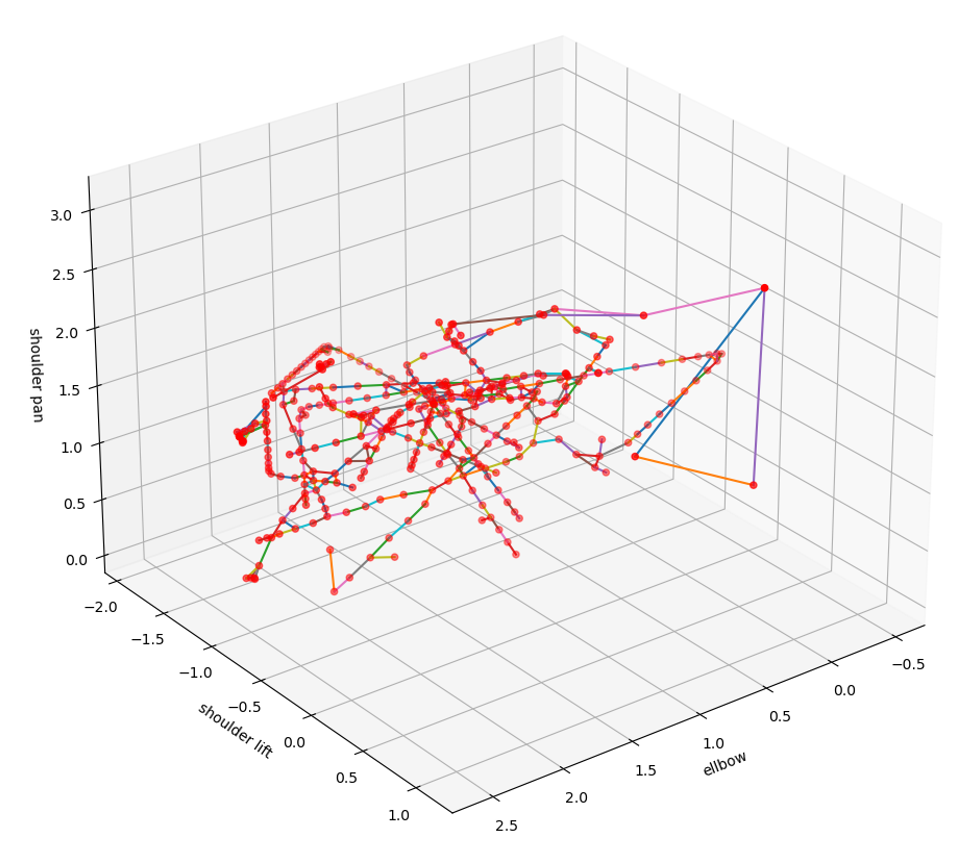}}
		\caption{\centering MGNG.}
	\end{subfigure}
	\begin{subfigure}[b]{0.3\textwidth}
		\centering
		\scalebox{1}[0.95]{\includegraphics[width=\textwidth]{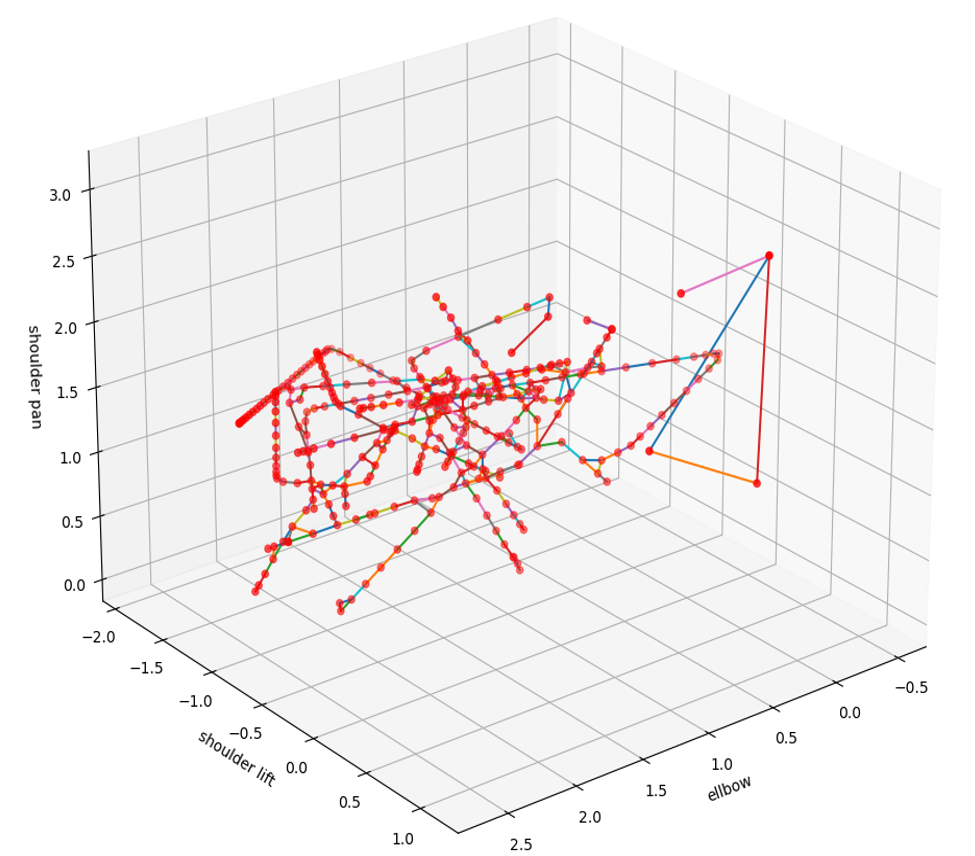}}
		\caption{\centering $\gamma$-GNG}
	\end{subfigure}
	\begin{subfigure}[b]{0.3\textwidth}
		\centering
		\scalebox{1}[0.95]{\includegraphics[width=\textwidth]{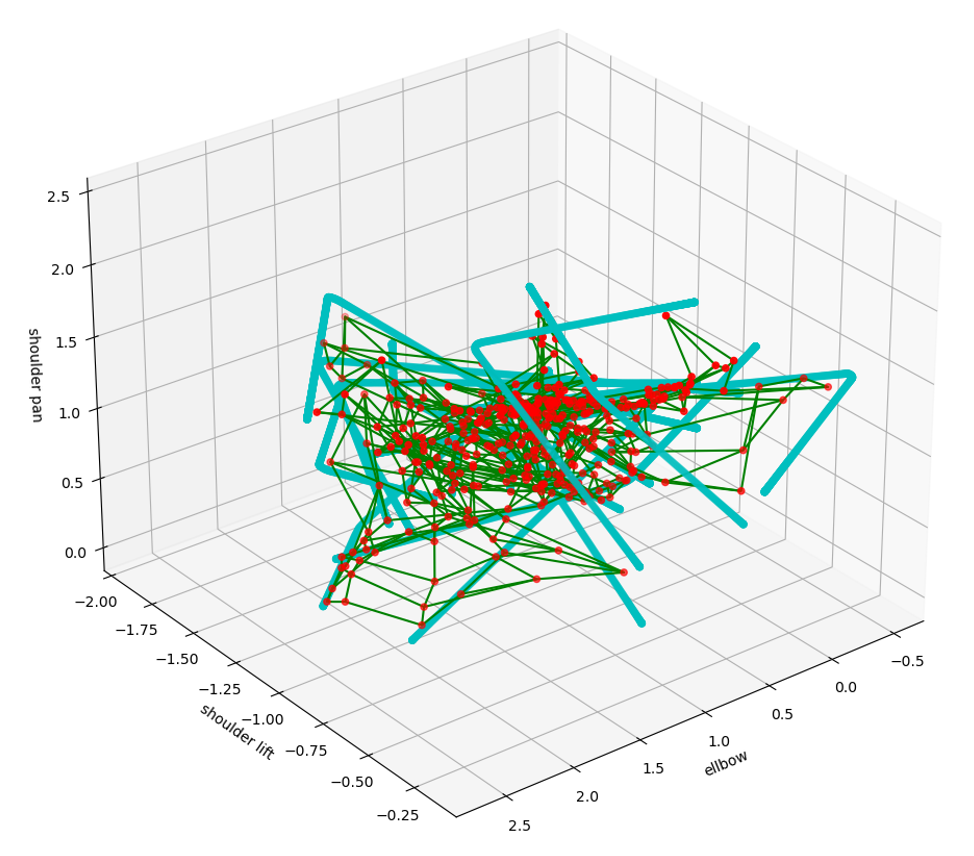}}
		\caption{\centering MSOM}
		\label{fig:vgl_mSOM}
	\end{subfigure}
	\begin{subfigure}[b]{0.3\textwidth}
		\centering
		\scalebox{1}[0.95]{\includegraphics[width=\textwidth]{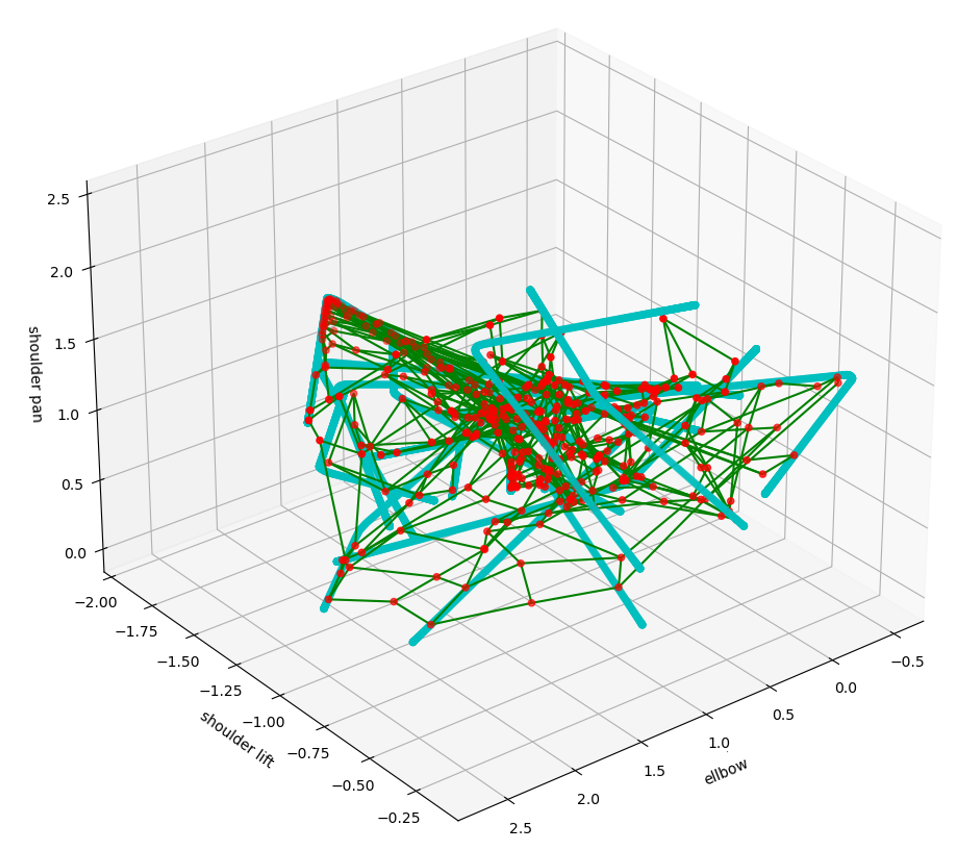}}
		\caption{\centering $\gamma$-SOM}
		\label{fig:vgl_gamma-SOM}
	\end{subfigure}
	\caption{SONN Comparison with 15 Trajectories. 
		The red scatter dots represent the neurons' weight vectors and the lines indicate the learned connections. For the SGNG green lines indicate the topological connections between segments while segments are illustrated by red lines between two neurons.}
	\label{fig:SONN_vgl_3Dinput}
\end{figure}
As postulated in hypothesis~\ref{hyp:2} it is assumed, that SONN models which consider the temporal context should be able to learn a better path preservation in comparison to pure GNGs. 
In \autoref{fig:SONN_vgl_3Dinput}, it is depicted that all GNG based models produce a proper estimation of the input space. The basic GNG showed similar or even better path preservation than all models with temporal context. The SGNG shows several splits along individual trajectories. 
Furthermore, all models generated connections between close neurons of two closely passing trajectories. Technically, these are wrong spatio-temporal connections, which was expected for the pure GNG model, but not for the models considering the temporal context. However, these connections between different trajectories at close points allow a smooth switch between different taught trajectories during the path planning.
In contrast to all GNG-based versions, the SOM based models perform significantly worse in terms of quantization and topological path preservation. 
Due to their topological mismatch the SOM-based models are somewhat folded in the higher dimensional space. This can be imagined like a carpet which tries to cover a 3D space and which is therefore folded and crumpled to generate some volume. Thereby, two carpet folds might be very close to each other in the 3D space but are far away from each other following the carpet course. This leads to long detours in the output space of the SOM models during the path finding (see \autoref{sec:path_analysis}). Furthermore, the lower dimensional structure of the SOMs is not well suited to map the topological structures of the higher dimensional input space. In \autoref{fig:vgl_mSOM} \& \subref{fig:vgl_gamma-SOM}, the SOM-based models are displayed with the interpolated input data to illustrate their mismatch between input samples and learned neuron weights and the misleading connections between the neurons estimating different input trajectories, resulting from the topological mismatch. Thus, the SOM based models build longer connections than the GNG based models, contrary to hypothesis 3.

\subsection{Parameter setting} \label{sec:parameter_setting}
For space reasons, the parameter analysis has been carried out representatively with GNG. Constant parameters are listed in \autoref{tab:constant_parameters}. A survey of SOM-specific quality measures is given in \cite{Forest2020}. In the following, the QE is supplemented by the C-measure (CM), which provides information about neighborhood preservation between spaces. In \autoref{tab:parametest_GNG}, the results and the specific settings for individual experiments are listed. Obviously, the more neurons a network has, the higher the number of connections. If relating column \textit{\#N}, stating the final size of the network, with \textit{QE} a correlation between an increased neuron number and an improved QE stands out. 
\begin{table}[h!]
	\begin{center}
		{\def\arraystretch{2}\tabcolsep=5pt
			\begin{tabular}{c c c c c c c} 
				\toprule
				\textbf{start size}  		& 	
				\textbf{$\eta_{BMU}$} 		& 	 	%
				\textbf{$\eta_n$} 		& 			%
				\textbf{$\lambda$} 		& 				%
				\textbf{$\zeta$} 		& 			%
				\textbf{$\delta$} 			& 	
				\textbf{$\epsilon$ } 		\\		%
				
				4        	& 			%
				0.06        & 			%
				0.005       & 			%
				20 			&			%
				0.995 		&  			%
				0.3 		&
				100 		\\			%
				\bottomrule
			\end{tabular}
		} %
		\caption{Constant parameters for the experiments regarding the parameter setting. $\eta_{BMU}$ is the factor for how strong the BMU is pulled towards the input vector. $\eta_n$ refers to how strong its neighbors are pulled towards the BMU. $\lambda$ defines the number of steps until a new neuron is inserted. $\zeta$ is the error counter and $\epsilon$ the max edge age.}
		\label{tab:constant_parameters}
	\end{center}
\end{table} 
However, as this is not a uniform ratio, it can be assumed that other factors besides the network size exist. The results show that blocking neurons, as done for column \textit{\#B}, worsens the QE, which is somewhat surprising, since in \cite{Steffen2021_dimreduction} it was assumed that it is bad if single neurons are selected as the BMU more often in succession. Additionally, it was observed that blocked neurons increase the number of synapses and also worsen the CM. The larger the CM, the further away topologically distant neurons are from each other in terms of weight. This means that in networks where BMUs are blocked, more connections exist between neurons further apart in terms of weight in the output space, which can lead to joint angle jumps in the subsequent path planning.
\begin{table}[h!]
	\begin{center}
		{\def\arraystretch{2}\tabcolsep=5pt
			\begin{tabular}{c c c c c c c c c c} 
				\toprule
				& \textbf{\#R} & \textbf{norm} & \textbf{\#B} &  \textbf{int.}  & \textbf{\#N} & \textbf{QE} & \makecell{\textbf{\# Con. /} \\ \textbf{CM}} & \makecell{\textbf{\# Red. /} \\ \textbf{CM}} \\
				\midrule
				1.1 & 1 & euclid 					& 0 & No & 2540  & 0.0852  & \makecell{7475 / \\ 10.18} 	& \makecell{6262 / \\ 12.80}  \\
				1.4 & 1 & \makecell{euclid \\ + cos} & 0 & No & 2540  & 0.0855 & \makecell{7515 / \\ 9.60} 		& \makecell{6343 / \\ 12.67}  \\
				1.6 & 1 & \makecell{euclid \\ + cos} & 2 & No & 2540  & 0.103  & \makecell{13967 / \\ 7.40} 	& \makecell{10194 / \\ 10.29}  \\     
				9.1 & 1 & \makecell{euclid \\ + cos} & 2 & 2x & 9965  & 0.0571 & \makecell{60990 / \\ 9.42}		& \makecell{55210 / \\ 10.75} \\
				9.2 & 4 & \makecell{euclid \\ + cos} & 2 & No & 10149 & 0.053  & \makecell{43902 / \\ 10.13} 	& \makecell{39661 / \\ 11.55} \\
				9.3 & 1 & \makecell{euclid \\ + cos} & 0 & 2x & 9964  & 0.0439 & \makecell{25479 / \\ 14.40}	& \makecell{24649 / \\ 15.09} \\
				9.4 & 4 & \makecell{euclid \\ + cos} & 0 & No & 10149 & 0.0436 & \makecell{24303 / \\ 13.90}	& \makecell{23310 / \\ 15.39} \\
				\bottomrule
			\end{tabular}
		} %
		\caption{Parameter study for GNG. \textit{\#R} is the number of training runs, thus, how often a net sees each sample. \textit{\#B} gives the number of blocked neurons. 2x Int. means that the input data was interpolated two times, thus the amount of input data was increased by the factor 4. \textit{\#N} refers to the number of neurons of the trained network. The two last columns contain two values, firstly, the amount of connections and secondly, the respective CM. \textit{\# Red.} refers to the amount of connections after the reduction applied for the Chebyshev distance.}
		\label{tab:parametest_GNG}
	\end{center}
\end{table} 
 The results of column \textit{\#R} (number of training runs) and \textit{int. input} suggest that two-fold linear interpolation, quadrupling the samples per trajectory, improves the QE more than more frequent (4-fold) learning of the same samples, despite almost the same final network size. Comparing tests 1.* and 9.*, the effect of longer training while constantly inserting
new neurons after a certain number of iterations becomes apparent. The QE decreases substantially while the CM increases with a larger network. \begin{figure}[hbt!]
	\centering
	\begin{subfigure}[b]{0.39\textwidth}
		\centering
		\includegraphics[width=\textwidth]{figures_paper/GNG}
		\caption{GNG (Test 9.4) \\ no neuron blocking}
		\label{subfig:coverage_14}
	\end{subfigure}
	\hfill
	\begin{subfigure}[b]{0.39\textwidth}
		\centering
		\includegraphics[width=\textwidth]{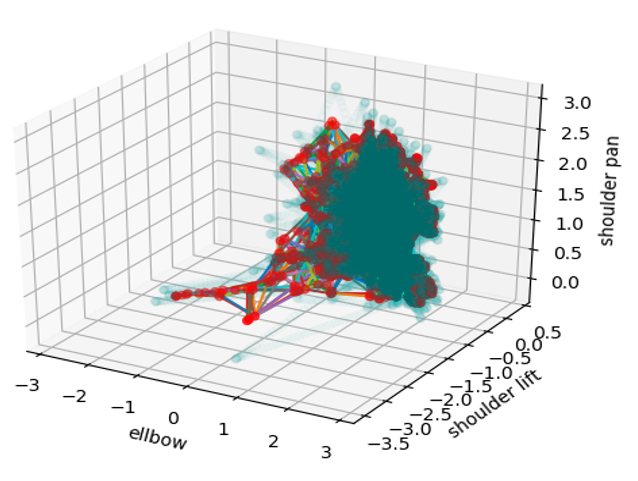}
		\caption{GNG (Test 9.2) \\ blocked neurons} %
		\label{subfig:coverage_16} 
	\end{subfigure}
	\caption{Coverage plots of the GNG output spaces displayed for 3 joint angles but learned with 6D input vectors.
		The blue scatter dots are the input samples while the red once represent the learned weight vectors of the neurons.
		The lines between red dots indicate learned connections.
	}
	\label{fig:1.4vs1.6}
\end{figure}
Regarding the CM, an increase of neurons and synapses does not automatically imply an improvement. This suggests that topological connections between distant neurons in the weight space still exist. One possible reason is that obsolete connections between neurons remain, which were connected before the network expanded. Another possible explanation is that the number of neurons is still too small to form a denser sampled paths and the neurons that are relatively far apart in the weight space are indeed still legitimately connected.
However, reducing connections, as done in the last column of \autoref{tab:parametest_GNG}, leads to a significant increase of the CM. As only connections between neurons are deleted, which also have an alternative indirect connection, this suggests that the constant CM for an increasing network size, is caused by too many old connections. 
As some of the synapses above the threshold always remain, it can be assumed that the neurons are either not optimally connected to each other, or there are outliers that are poorly connected to the other neurons. \\
Regarding the paths, generated with the different versions, the reduced connections lead to more neurons on the paths and thus, to higher path resolutions (see \autoref{fig:pa_GNG}). This correlates with the increasing CM. A similar correlation can be seen for bigger networks with more neurons. These larger networks imply higher path resolutions and a higher CM.
\autoref{fig:1.4vs1.6} shows plots regarding the coverage for experiments listed in \autoref{tab:parametest_GNG}. 
The GNG of test 9.4 in \autoref{subfig:coverage_14} shows a good coverage of the input space. In contrast, the neurons of test 9.2 in \autoref{subfig:coverage_16} are drawn towards the center, the input space is not covered as well and the  QE is worse. This also discourages hypothesis~\ref{hyp:1}, as blocking of BMUs for few iterations to enforce a more diverse BMU selection even worsened the results.

\subsection{Comparison of the Different Models} \label{sec:sonn_comparison}
\begin{figure}[hbt!]
	\centering
	\begin{subfigure}[b]{0.3\textwidth}
		\centering
		\includegraphics[width=\textwidth]{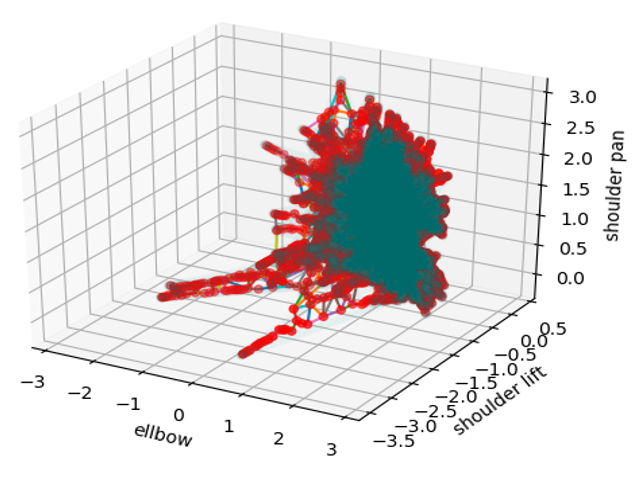}
		\caption{GNG} %
		\label{subfig:GNG_coverage_94}
	\end{subfigure}
	\hfill
	\begin{subfigure}[b]{0.3\textwidth}
		\centering
		\includegraphics[width=\textwidth]{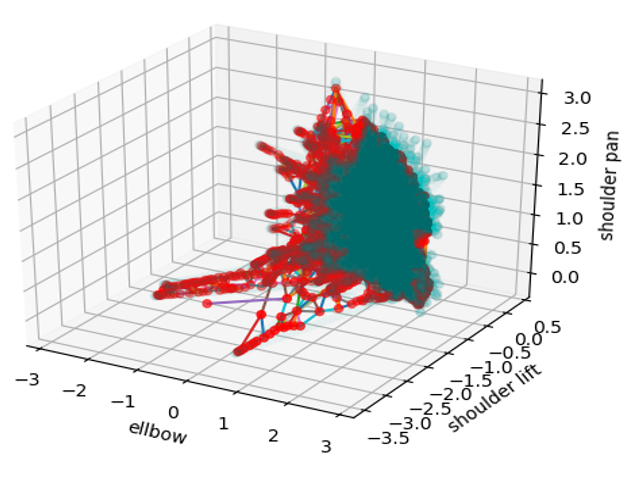}
		\caption{MGNG} %
		\label{subfig:MGNG_coverage_21}
	\end{subfigure}
	\hfill
	\begin{subfigure}[b]{0.3\textwidth}
		\centering
		\includegraphics[width=\textwidth]{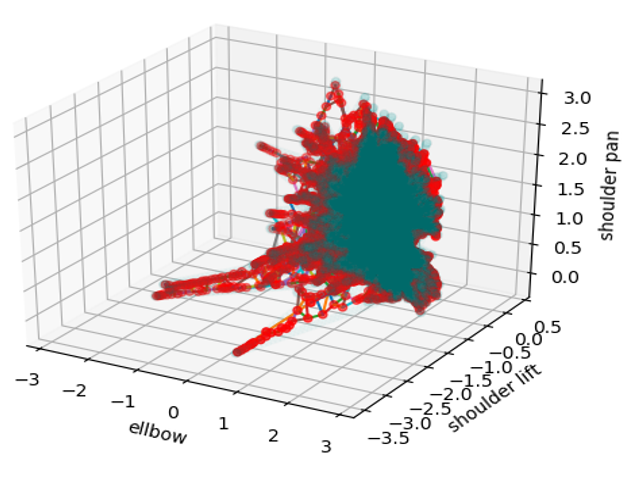}
		\caption{$\gamma$-GNG} %
		\label{subfig:GGNG_coverage_21}
	\end{subfigure}
	\\
	\begin{subfigure}[b]{0.3\textwidth}
		\centering
		\includegraphics[width=\textwidth]{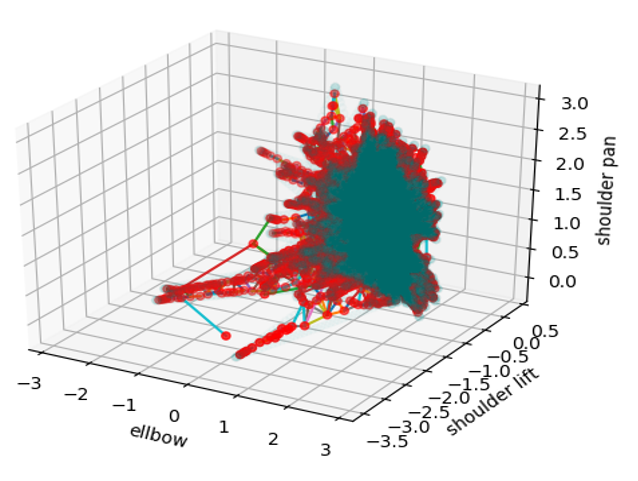}
		\caption{SGNG} %
		\label{subfig:SGNG_coverage_21}
	\end{subfigure}
	\hfill
	\begin{subfigure}[b]{0.3\textwidth}
		\centering
		\includegraphics[width=\textwidth]{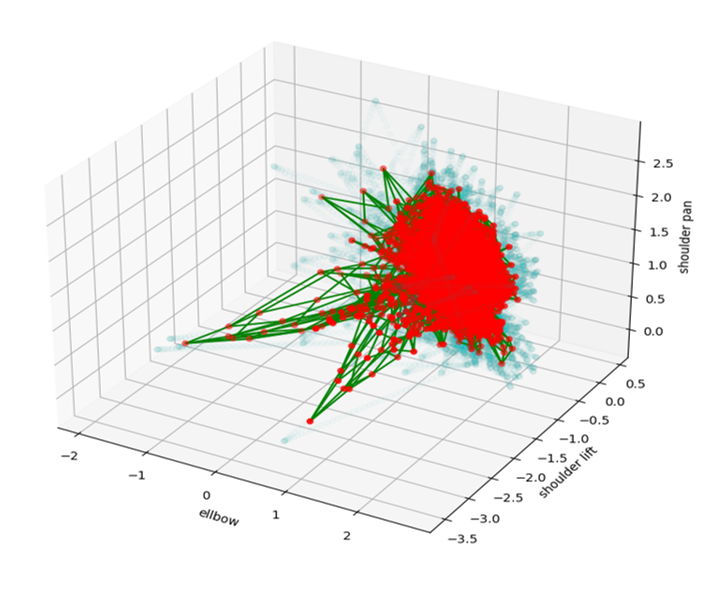}
		\caption{MSOM} %
		\label{subfig:MSOM_coverage_21} 
	\end{subfigure}
	\hfill
	\begin{subfigure}[b]{0.3\textwidth}
		\centering
		\includegraphics[width=\textwidth]{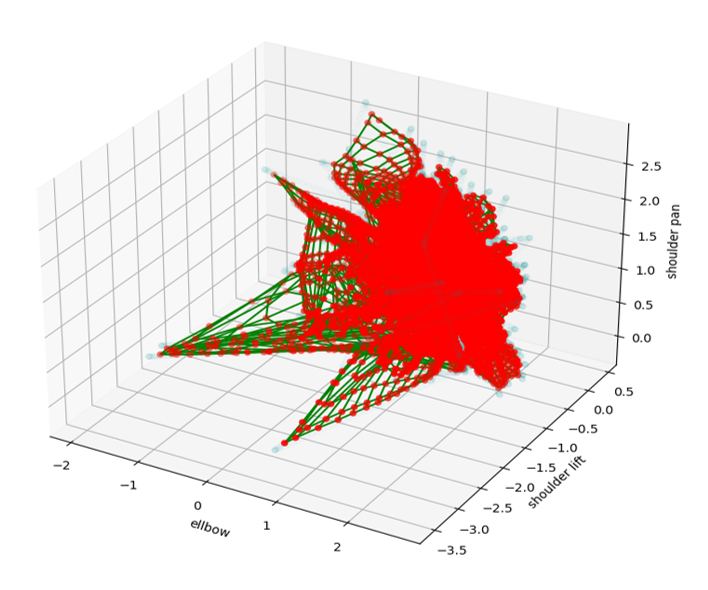}
		\caption{$\gamma$-SOM} %
		\label{subfig:GSOM_coverage_21} 
	\end{subfigure}
	\hfill
	\caption{Plots of the learned output spaces for the different SONN models. The red scatter dots represent the learned weight vectors and the lines between the red dots indicate the learned connections between the neurons. Except for the SOM based models which are connected to their topological neighbors. For parameters of GNG-based networks see \autoref{tab:constant_parameters} and \autoref{tab:parametest_GNG}, for SOM-based models \autoref{tab:som_params}.}
	\label{fig:sonn_converage}
\end{figure}

Regarding all GNG-based models, $\gamma$-GNG, MGNG, SGNG, the parameters are from test 9.4 (see \autoref{tab:parametest_GNG}), as it generated the best trajectories. Hence, the GNG-based models of \autoref{tab:sonn_comparison_gng} had 4 training runs, no blocked neurons or interpolation.
\begin{table}[h!]
	\begin{center}
		{\def\arraystretch{2}\tabcolsep=5pt
			\begin{tabular}{c c c c c c c} 
				\toprule
				\textbf{\#R} 	& \textbf{\#N} 	& \textbf{dim}  & \textbf{$\sigma$} & \textbf{$\eta$} & \textbf{$\alpha$} & \textbf{$\beta$}\\
				\midrule
				4 				& 	100x100 (10000)		& 2D	& 5.0  & 0.2 	& 0.3 & 0.7 \\
				\bottomrule
			\end{tabular}
		} %
		\caption{Parameter setting for the SOM-based models. The output dimension is defined by \textit{dim}, $\sigma$ is the neighborhood rate and $\eta$ the learning rate. }
		\label{tab:som_params}
	\end{center}
\end{table}The parameters for the SOM-based models, stated in \autoref{tab:som_params}, were selected to maximize comparability. Respectively, the input data was not interpolated either.
It is noticeable in \autoref{tab:sonn_comparison_gng} and \autoref{fig:sonn_converage} that coverage and QE are closely related. GNG, $\gamma$-GNG and SGNG have the best QE and also the best coverage of the input space. On the other hand the MGNG and especially the MSOM are clustered towards the center and obtain therefore a substantially worse QE. Therefore, hypothesis~\ref{hyp:4} is proven as correct. 
The biggest difference, however, seems to be in the connections. The MGNG, while having the worst CM, has established twice as many connections as any other GNG-based model. This indicates many long connections and therefore, large joint angle jumps and a rather bad path resolution. Referring to the coverage plots in \autoref{fig:sonn_converage}, the GNG and the $\gamma$-GNG seem to have the best topology preserving connectivity between the neurons, while other types show more long connections and outliers between different individual input trajectories. The SGNG contains substantially less neurons than other GNG based models. Apparently, many established connections and inserted neurons were deleted again during the training process.
\begin{table}[h!]
	\begin{center}
		{\def\arraystretch{2}\tabcolsep=5pt
			\begin{tabular}{c c c c c} 
				\toprule
				\textbf{type} & \textbf{\#N} & \textbf{QE} & \textbf{\# Con. / CM} & \textbf{\# Red. / CM}  \\
				\midrule
				\textbf{GNG} 			& 10149	& 0.0436 & 24303 / 13.9 	& 23310 / 15.38  \\
				\textbf{MGNG} 			& 10149 & 0.0623 & 54571 / 7.25	& 46608 / 9.88  \\			
				\textbf{$\gamma$-GNG}  	& 10149 & 0.0458 & 29006 / 12.34	& 27902 / 13.75 \\
				\textbf{SGNG} 			& 8968 	& 0.0501 & 24589 / 8.973	& 22884 / 12.29 \\
				\textbf{MSOM} 			& 10000 & 0.1149 & 19800 / 61.006  	& -- \\
				\textbf{$\gamma$-SOM} 	& 10000 & 0.0575 & 19800 / 61.737 					& -- \\
				\bottomrule
			\end{tabular}
		} %
		\caption{SONN comparison regarding their QE and CM. \#N is the size of the trained networks, \#N Con. the number of connections and \#N Red. the number of reduced connections. As SOMs have a static number of synapses the reduction is only applied for GNG-based models.}
		\label{tab:sonn_comparison_gng}
	\end{center}
\end{table} 
SOM based models show a substantially worse topology preservation compared to GNG based models. Firstly, many neurons lie between different input trajectories and the connections are linking wide distances between different input trajectories. Secondly, the mismatch between the topological dimensions becomes evidently and the "folding" of the 2D shaped SOMs in the 3D space can be clearly seen. 
Note, however, that the actual input space is 6D for which the topological mismatch becomes an even bigger problem. This also explains the MSOM's huge CM. Due to the folding effect, the paths in the output space can become ridiculously long for samples which are actually close in the input space.

\subsection{Path Analysis} \label{sec:path_analysis}
\begin{figure}[ht!]
	\centering
	\begin{subfigure}[b]{0.32\textwidth}
		\centering
		\scalebox{1}[1]{\includegraphics[width=\textwidth]{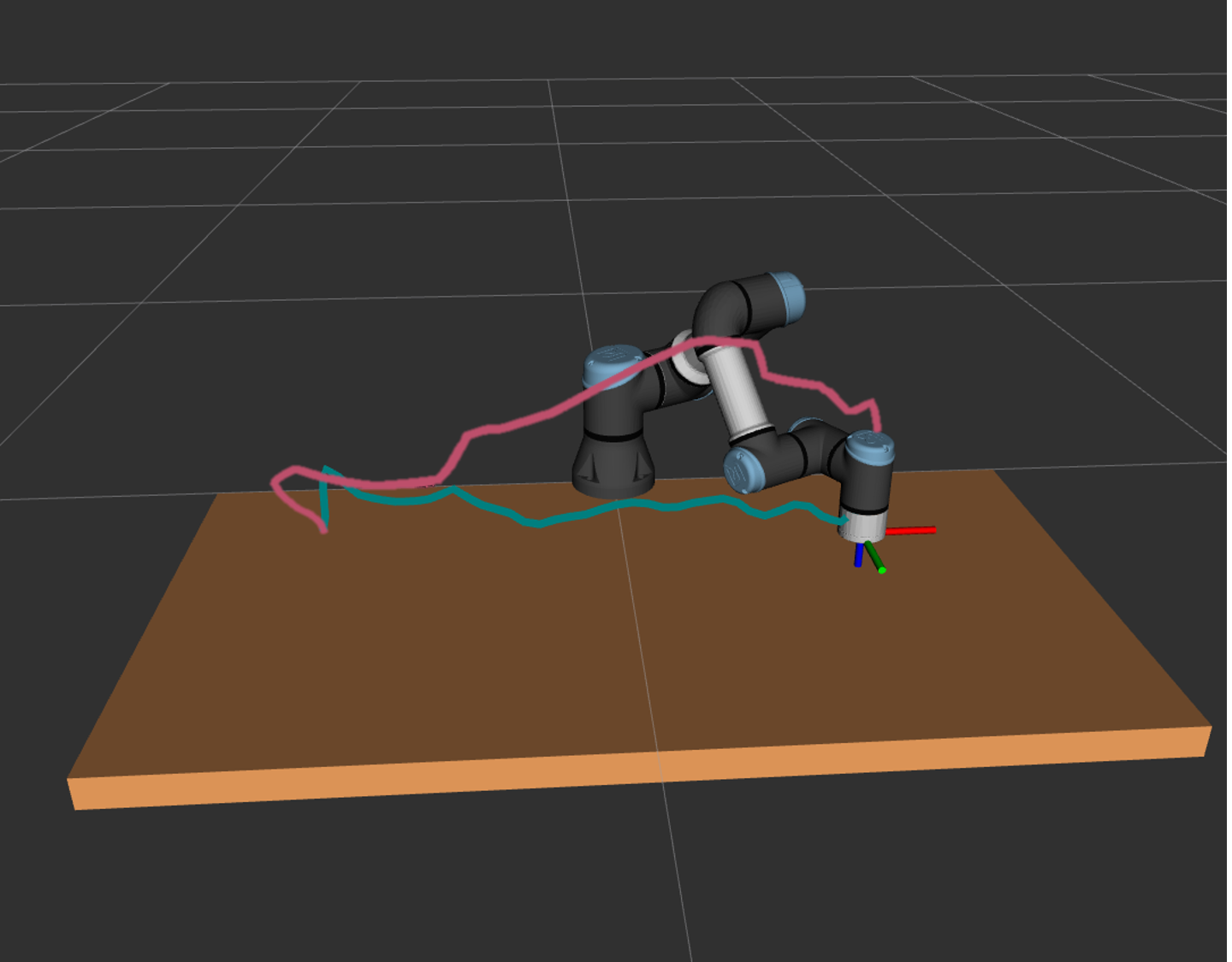}}
		\caption{\centering GNG}
		\label{fig:pa_GNG}
	\end{subfigure}
	\begin{subfigure}[b]{0.32\textwidth}
		\centering
		\scalebox{1}[1]{\includegraphics[width=\textwidth]{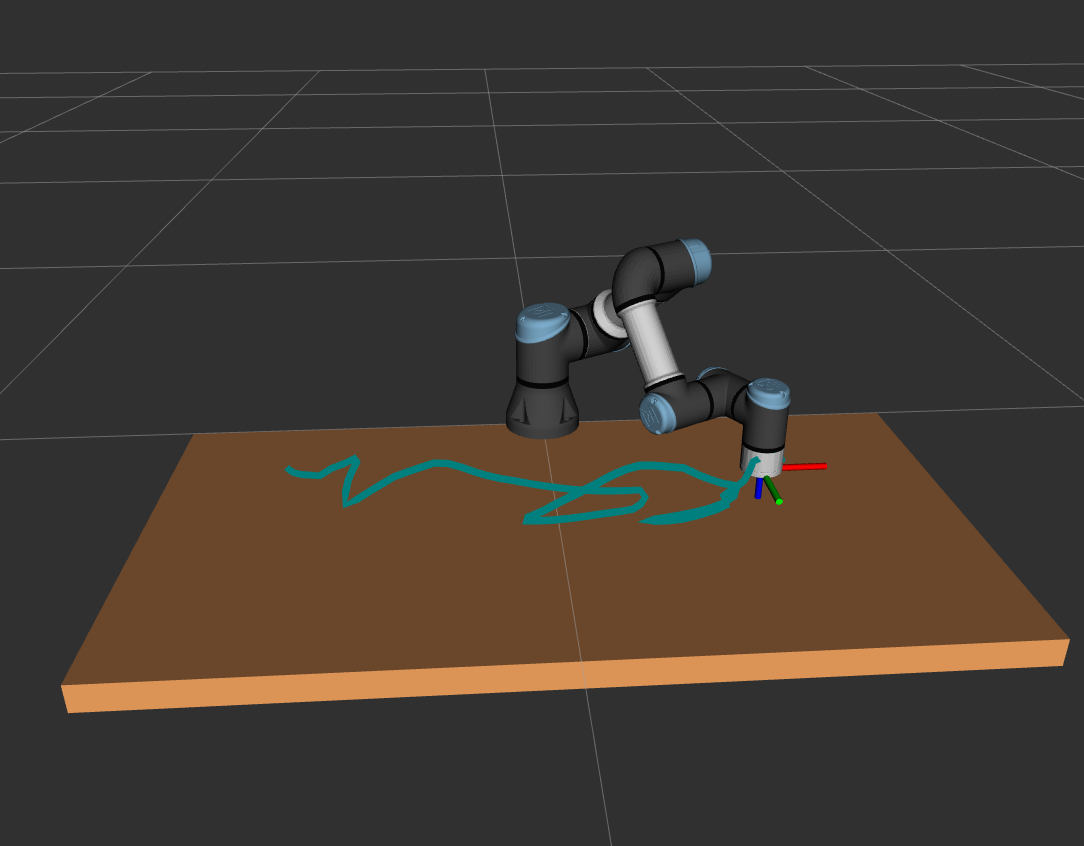}}
		\caption{\centering MSOM}
		\label{fig:pa_MSOM}
	\end{subfigure}
	\begin{subfigure}[b]{0.32\textwidth}
		\centering
		\scalebox{1}[1]{\includegraphics[width=\textwidth]{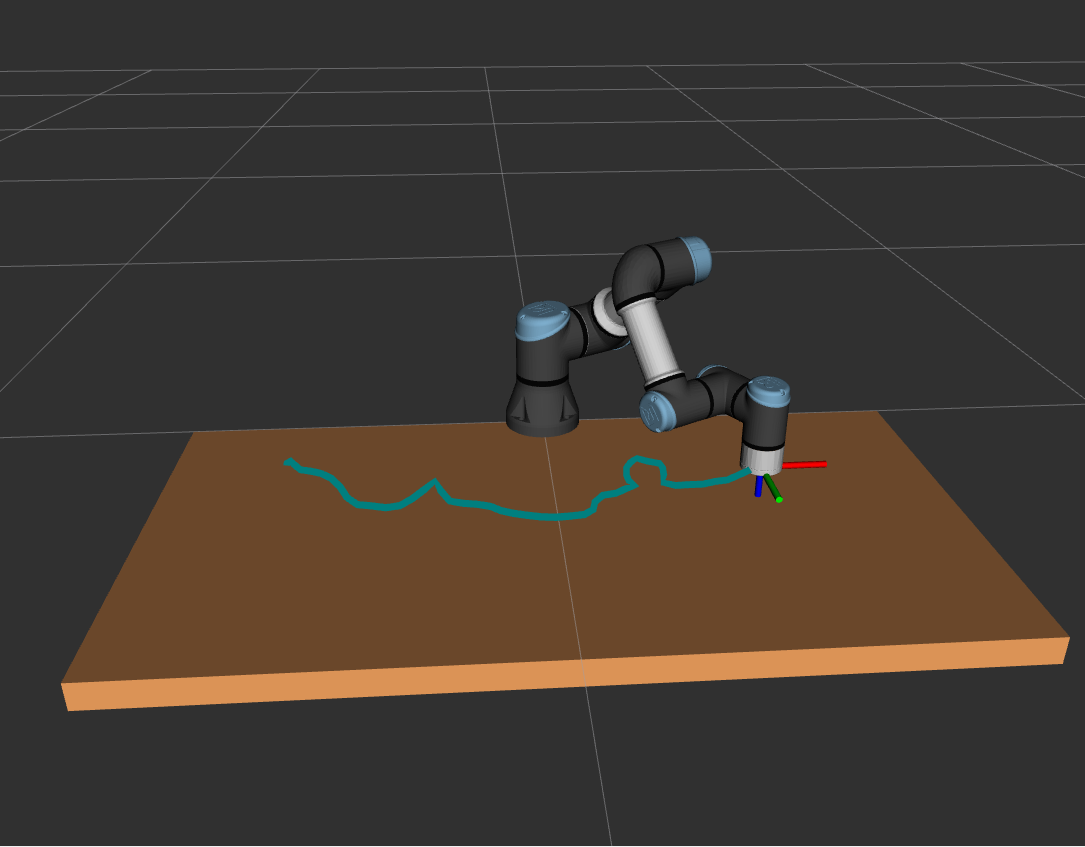}}
		\caption{\centering $\gamma$-SOM}
		\label{fig:pa_gammaSOM}
	\end{subfigure}
	\caption{Comparison of generated motions with identical start and target configuration for different SONN models. The blue curve illustrates the Cartesian 3D trajectory of the endeffector. In \protect\subref{fig:pa_GNG} an additional comparison between the path with original and reduced connections, regarding the GNG is provided.}
	\label{fig:SONN_path_vgl_without_obst}
\end{figure}

To allow a fair analysis of the resulting path, exemplary generated motions in the obstacle-free task space with the reduced C-spaces were used, as depicted in \autoref{fig:SONN_path_vgl_without_obst}. The graphic shows only the endeffector's path for GNG, MSOM and $\gamma$-SOM, as all GNG-based networks produced similar results. 
From all GNG-based models, the MGNG achieved the worst path resolution. Which fits to the GNG's and $\gamma$-GNG's higher CM in comparison to the MGNG.
Additionally, for the GNG a comparison between original and reduced connections, as introduced in \autoref{sec:connection_recudtion}, is displayed. 
Planning with the original longer connections results in a lower path resolution (27 vs. 32 neurons on the paths in \autoref{fig:pa_GNG}). Furthermore, the reduced connections seem to lead to more direct trajectories, while the original connections often produce wide-ranging detours.
The twists and loops in the paths of the SOM-based models in \autoref{fig:pa_MSOM} and \autoref{fig:pa_gammaSOM} are not surprising regarding the results from the previous SONN analysis in \autoref{sec:sonn_comparison}. Due to the topological mismatch between the 2D shape of the SOMs’ output shape and the 6D input space, the SOMs are somewhat "folded". Hence, topologically neighbored neurons represent states which are also topologically neighbored in the input space. However, joint states which are close in the input space are not necessarily nearby located in the SOMs' output space. Thus, trajectories generated by SOM-based models contain many neurons and are relatively smooth with quite high path resolutions and few joint angle jumps but show great detours and loops. Thereby, the $\gamma$-SOM outperforms the MSOM. 
Obviously, the many detours and loops of the trajectories generated by all  SOM-based models are not practical. The reason for ridiculous long paths of the SOM based models lies especially in the topological mismatch between the 2D output space of the SOMs and the 6D input space. \\
The previous qualitative and quantitative analysis have shown that also the GNG-based models contain few long connections which can lead to joint angle jumps. However, they produce much less large joint angle jumps than SOM-based models, due to their flexible fitting to higher dimensional input spaces.
Hence, the reason for shorter paths of the GNG-based in comparison to the SOM-based models, as it was observed in \cite{Steffen2021_dimreduction}, is not that the GNGs produce many long connections with large joint angle jumps, as postulated in hypothesis 3. Actually, SOM-based models produce ridiculously long paths along their folded structures and GNG-based once show much more feasible and reasonable paths.

\section{Conclusion} \label{sec:conclusion}
In this work, we extended and adapted the approach presented in \cite{Steffen2021_dimreduction}. 
As training data, 6 DOF robot trajectories are used instead of 7 DOF human motions. This is important for applying our approach to a real applications and proves that a transfer of the method to another robot kinematics is possible. As the core part of this work, we conducted an evaluation about different SONN models for reducing the C-space. 
 Based on their characteristics, we excluded 3 network models while maintaining the MSOM \cite{Strickert2005}, GNG \cite{Fritzke1995} and MGNG \cite{Andreakis2009}. Respectively the SGNG \cite{Vergara2016}, $\gamma$-SOM model \cite{Estevez2009} and $\gamma$-GNG \cite{Estevez2013} were introduced. 
Four hypotheses were formulated based on previous results as well as findings in literature. These were tested with a qualitative and quantitative analysis. The most important findings are that all SONN types reproduce the C-space properly but the best coverage was achieved by the GNG \cite{Fritzke1995} and the $\gamma$-SOM model \cite{Estevez2009}. 
SOM- and GNG-based networks are equivalent in terms of coverage. However, SOM based models show a substantially worse topology preservation than GNG based models. Overall, the GNG and the $\gamma$-GNG showed the best results, while the CM values of the GNG seemed slightly superior. Furthermore, the deletion of extra long connections could improve the trajectory quality and reduce long joint angle jumps on the path.
It was shown in detail that SOMs lie folded in the multidimensional space, due to their 'topological mismatch' and therefore, not all adjacent points in the original C-space are also automatically adjacent in the reduced C-space.
Based on our evaluation, GNG and $\gamma$--GNG are recommended for future projects with the presented approach.
Additionally, it became apparent that the ratio between the number of neurons and samples plays an important role. While too few neurons lead to a poor resolution and coverage, too sparsely sampled data in combination with many neurons leads to splits in the graph structure of the output space of GNG based models. \\
For future work, the system shall take static, and most importantly, also dynamic obstacles into account. Therefore a very fast and efficient transformation of obstacles in the Cartesian task
space to the C-space is required. Finally, execution on a robotic demonstrator is planned, whereby integrating a vision component is necessary.

\section*{Acknowledgment}
This research has been supported by the European Union's Horizon 2020 Framework Programme for Research and Innovation under the Specific Grant Agreement No. 945539 (Human Brain Project SGA3).

\bibliographystyle{splncs03} 	%
\bibliography{library,evaluation}

\begin{thebibliography}{10}
\providecommand{\url}[1]{\texttt{#1}}
\providecommand{\urlprefix}{URL }

\bibitem{Andreakis2009}
Andreakis, A., Hoyningen-Huene, N.V., Beetz, M.: {Incremental unsupervised time
  series analysis using merge growing neural gas}. Lecture Notes in Computer
  Science  5629 LNCS,  10--18 (2009)

\bibitem{Barreto2003}
Barreto, G.d.A., Ara{\'{u}}jo, A.F.R., Ritter, H.J.: {Self-Organizing Feature
  Maps for Modeling and Control of Robotic Manipulators}. J. Intell. Robot.
  Syst  36(4),  407--450 (apr 2003)

\bibitem{Bishop1997}
Bishop, C.M., Hinton, G.E., Strachan, I.G.D.: {GTM through time — Aston
  Research Explorer}. ICANN pp. 111--116 (1997)

\bibitem{Chappell1993}
Chappell, G.J., Taylor, J.G.: {The temporal Koh{\o}nen map}. Neural Networks
  6(3),  441--445 (jan 1993)

\bibitem{Cottrell2016}
Cottrell, M., Olteanu, M., Rossi, F., Vialaneix, N.: {Theoretical and applied
  aspects of the self-organizing maps}. WSOM  11 (2016)

\bibitem{Estevez2009}
Est{\'{e}}vez, P.A., Hern{\'{a}}ndez, R.: {Gamma SOM for temporal sequence
  processing}. Lecture Notes in Computer Science  5629 LNCS,  63--71 (2009)

\bibitem{Estevez2013}
Est{\'{e}}vez, P.A., Vergara, J.R.: {Nonlinear Time Series Analysis by Using
  Gamma Growing Neural Gas}. Adv. in Intelligent Systems and Computing  198,
  205--214 (2013)

\bibitem{Euliano1999}
Euliano, N.R., Principe, J.C.: {A Spatio-Temporal Memory Based on SOMs with
  Activity Diffusion}. In: Kohonen Maps, pp. 253--265. Elsevier (jan 1999)

\bibitem{Fontinele2016}
Fontinele, H.I., Melo, D.B., Barreto, G.A.: {Local models for learning inverse
  kinematics of redundant robots: A performance comparison}. In: Adv. in Intel.
  Sys. and Computing. vol. 428, pp. 177--187. Springer Verlag (2016)

\bibitem{Forest2020}
Forest, F., Lebbah, M., Azzag, H., Lacaille, J.: {A Survey and Implementation
  of Performance Metrics for Self-Organized Maps}. CoRR  (nov 2020)

\bibitem{Fritzke1995}
Fritzke, B.: {A Growing Neural Gas Network Learns Topologies}. Adv. in Neural
  Information Processing Systems  (1995)

\bibitem{Hagenbuchner2003}
Hagenbuchner, M., Sperduti, A., Tsoi, A.C.: {A self-organizing map for adaptive
  processing of structured data}. Trans. on ANN  14(3),  491--505 (may 2003)

\bibitem{Hammer2005}
Hammer, B., Micheli, A., Neubauer, N., Sperduti, A., Strickert, M.: {Self
  organizing maps for time series}. WSOM pp. 115--122. (2005)

\bibitem{Kohonen1982}
Kohonen, T.: {Self-organized formation of topologically correct feature maps}.
  Biological Cybernetics  43(1),  59--69 (jan 1982)

\bibitem{Kohonen1997}
Kohonen, T.: {Self-Organizing Maps} (1997)

\bibitem{Kohonen2013}
Kohonen, T.: {Essentials of the self-organizing map}. Neural Networks  37,
  52--65 (jan 2013)

\bibitem{Koskela1998}
Koskela, T., Varsta, M., Heikkonen, J., Kaski, K.: {Temporal sequence
  processing using recurrent SOM}. KES  1 (1998)

\bibitem{Martinetz1991}
Martinetz, T., Schulten, K.: {A Neural-Gas Network Learns Topologies} (1991)

\bibitem{Martinetz1993}
Martinetz, T.: {Competitive Hebbian Learning Rule Forms Perfectly Topology
  Preserving Maps}. In: ICANN. Springer London (1993)

\bibitem{Martinetz1990}
Martinetz, T., Ritter, H.J., Schulten, K.J.: {Learning of Visuomotor
  Coordination of a Robot Arm with Redundant Degrees of Freedom}. Parallel
  Processing in Neural Systems and Computers  (1990)

\bibitem{Martinetz1994}
Martinetz, T., Schulten, K.: Topology representing networks. Neural Networks
  7(3),  507--522 (1994)

\bibitem{Miljkovic2017}
Miljkovic, D.: {Brief review of self-organizing maps}. In: MIPRO. pp.
  1061--1066. Institute of Electrical and Electronics Engineers Inc. (jul 2017)

\bibitem{Naumov2015}
Naumov, V., Karova, M., Zhelyazkov, D., Todorova, M., Penev, I., Nikolov, V.,
  Petkov, V.: {Robot path planning algorithm}. Int. J. Comput. Commun.  9
  (2015)

\bibitem{Prabhu1996}
Prabhu, S.M., Garg, D.P.: {Artificial neural network based robot control: An
  overview}. J. Intell. Robot. Syst  15(4),  333--365 (1996)

\bibitem{Saxon1990}
Saxon, J.B., Mukerjee, A.: {Learning the motion map of a robot arm with neural
  networks}. IJCNN pp. 777--782 (1990)

\bibitem{Schulz2004}
Schulz, R., Reggia, J.A.: {Temporally Asymmetric Learning Supports Sequence
  Processing in Multi-Winner Self-Organizing Maps}. Neural Computation  16(3),
  535--561 (mar 2004)

\bibitem{Simon2003}
Simon, G., Lendasse, A., Cottrell, M., Fort, J.C., Verleysen, M.: {Double SOM
  for long-term time series prediction}. WSOM pp. 35--40 (2003)

\bibitem{Somervuo2004}
Somervuo, P.J.: Online algorithm for the self-organizing map of symbol strings
  (oct 2004)

\bibitem{Steffen2021_dimreduction}
Steffen, L., Glueck, K., Ulbrich, S., Roennau, A., Dillmann, D.: {Reducing the
  Dimension of the Configuration Space with Self Organizing Neural Networks}.
  ICARM  (2021)

\bibitem{Strickert2005}
Strickert, M., Hammer, B.: {Merge SOM for temporal data}. Neurocomputing  64,
  39--71 (2005)

\bibitem{Terlemez2015}
Terlemez, {\"{O}}., Ulbrich, S., Mandery, C., Do, M., Vahrenkamp, N., Asfour,
  T.: {Master Motor Map (MMM) - Framework and toolkit for capturing,
  representing, and reproducing human motion on humanoid robots}. Int. Conf.
  Humanoid Robots pp. 894--901 (feb 2015)

\bibitem{Tino2004}
Tino, P., Kab{\'{a}}n, A., Kab{\'{a}}n, K., Sun, Y.: {A Generative
  Probabilistic Approach to Visualizing Sets of Symbolic Sequences}. Int. conf.
  on Knowledge discovery and data mining  4,  701--706 (2004)

\bibitem{VanHulle2012}
{Van Hulle}, M.M.: {Self-Organizing Maps}. Tech. rep. (2012)

\bibitem{Vergara2017}
Vergara, J.R., Est{\'{e}}vez, P.A.: {A strategy for time series prediction
  using segment growing neural gas}. WSOM  (aug 2017)

\bibitem{Vergara2016}
Vergara, J.R., Est{\'{e}}vez, P.A., Serrano, {\'{A}}.: {Segment growing neural
  gas for nonlinear time series analysis}. In: Adv. in Intelligent Systems and
  Computing. vol. 428, pp. 107--117. Springer Verlag (2016)

\bibitem{Villmann1997}
Villmann, T., Der, R., Herrmann, M., Martinetz, T.M.: {Topology preservation in
  self-organizing feature maps: Exact definition and measurement} (1997)

\bibitem{Voegtlin2001}
Voegtlin, T., Dominey, P.F.: {Recursive Self-Organizing Maps}. In: Adv. in SOM,
  pp. 210--215. Springer London (2001)

\bibitem{Wiemer2003}
Wiemer, J.C.: {The Time-Organized Map Algorithm: Extending the Self-Organizing
  Map to Spatiotemporal Signals}. Neural Computation  15,  1143--1171 (2003)

\end{thebibliography}

\end{document}